\documentclass[10pt,twocolumn,letterpaper]{article}

\usepackage{cvpr}      

\usepackage{graphicx}
\usepackage{amsmath}
\usepackage{amssymb}
\usepackage{booktabs}

\usepackage[pagebackref,breaklinks,colorlinks]{hyperref}

\usepackage{amsmath}
\usepackage{comment}
\usepackage{enumitem}
\usepackage{subcaption}
\usepackage{multirow}
\usepackage{makecell}
\usepackage{stfloats} 

\DeclareMathOperator*{\argmax}{arg\,max}

\usepackage[capitalize]{cleveref}
\crefname{section}{Sec.}{Secs.}
\Crefname{section}{Section}{Sections}
\Crefname{table}{Table}{Tables}
\crefname{table}{Tab.}{Tabs.}

\def\cvprPaperID{11982} 
\def\confName{CVPR}
\def\confYear{2023}

\title{Prompt Tuning for Zero-shot Compositional Learning}
\date{November 2022}
\author{Lingyu Zhang,
\enspace Ting Hua,
  \enspace Yilin Shen, 
  \enspace Hongxia Jin\\
  Samsung Research America \\
  {\{lingyu.zhang, ting.hua, yilin.shen, hongxia.jin\}@samsung.com}}
\begin{document}

\maketitle

\begin{abstract}
Open World Compositional Zero-Shot Learning (OW-CZSL) is known to be an extremely challenging task, which aims to recognize unseen compositions formed from seen attributes and objects without any prior assumption of the output space. 
In order to achieve this goal,  a model has to be ``smart'' and ``knowledgeable''.
To be smart, a model should be good at  reasoning the interactions between attributes and objects from the seen compositions.
While ``knowledgeable'' means the model owns ``common sense'' to the open world that can ``foresee'' some features of the unseen compositions. 
Most previous work focus on the ``smart'' part, while few of them provided an effective solution to achieve the  ``knowledgeable'' goal. 
In this paper, we proposed a framework named Multi-Modal
Prompt Tuning (MMPT) to inherit the ``knowledgeable'' property from the large pre-trained vision-language model. 
Extensive experiments show that our proposed MMPT obtains new state-of-the-art results in OW-CZSL task.
On the UT-Zappos dataset, MMPT pushes the AUC score to $29.8$, while the previous best score is $26.5$.
On the more challenging MIT-States dataset, the AUC score of MMPT is 1.5 times better than the current state-of-the-art.   

\end{abstract}

\section{Introduction}

Compositional generalization ability is a fundamental
trait of humankind\cite{lake2018generalization,lake2015human},  
allowing us to understand novel concepts through composing learned knowledge.
Compositional Zero-Shot Learning (CZSL) aims to simulate this human intelligence by learning a small amount of seen compositions and generalizing the recognition ability to unseen compositions. 
In the CZSL setting, one composition is a pair of object and attribute. 
Taking the image ``Old Elephant''   for example,
``Old'' is the attribute of object ``Elephant''.
At the testing time, the CZSL models are supposed to recognize the compositions never seen before by decomposing the known compositions in the training data (e.g., understand the concept of ``Old Truck'' through learning ``Old Elephant'' and ``New Truck'').
This is a very challenging task, as the different combinations of the same attribute and object will generate images with visual features various in shapes, colors, and textures.
For example, the two attributes of the same object can be fundamentally different in visual features (e.g., the ``Raw chicken'' and ``Sliced chicken'' ).
Similarly, one attribute will display completely differently on two objects (e.g., ``Old Truck'' and ``Old Elephant'').
 
  Previous CZSL models show that there is a chance to recognize the unseen pairs.
 However, their successes are only limited to the closed output space, where the prior knowledge of the unseen composition is assumed in the testing data. 
 Significant performance drops are observed when applying these methods to the Open World CZSL (OW-CZSL) task \cite{karthik2022kg-sp,mancini2021open}, which is a more realistic setting without limitation in the output search space. 
 
 \begin{figure*} [htbp]
\flushright
\includegraphics[width=\textwidth]{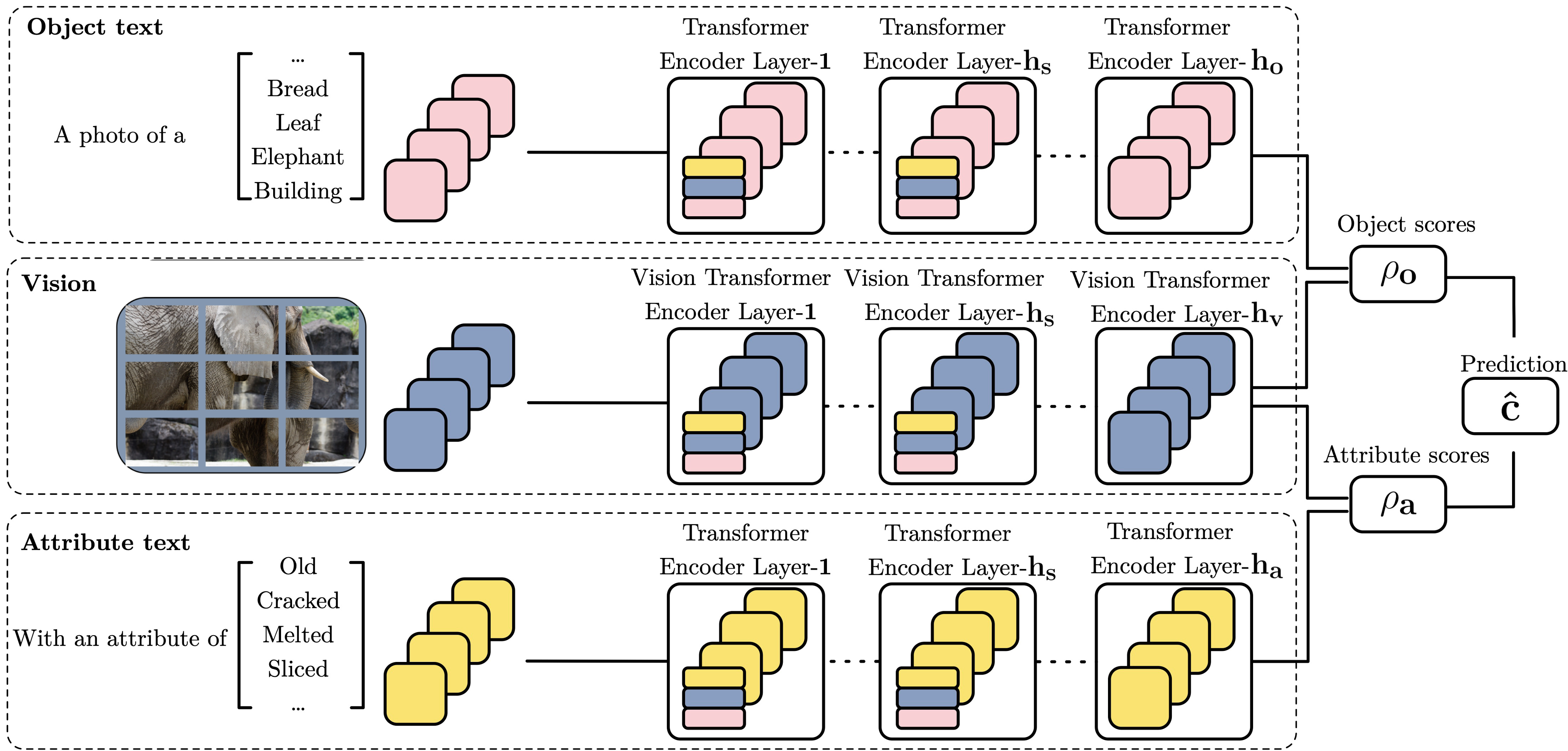}
\caption{The overview of our proposed MMPT. 
We construct three branches to process text and visual features: attribute branch, vision branch, and object branch.
Each branch has its own learnable parameters which are denoted as squares colored with pink (attribute branch), blue (vision branch), and yellow (object branch) .
The multicolored square represents the shared prompt across different branches. 
}
\label{fig:framework}
\end{figure*}

 What is the secret behind human's compositional learning ability? To answer this question,  imagining Tarzan who grows up  in the African jungle, can he recognize the ``Old Truck'' merely through looking at the images of ``Old Elephant'' and ``New Truck''? The answer is likely to be ``no''. 
 Although Tarzan is mankind, he has never seen a car before.
 It is unlikely for him to know the concept of ``Truck'' through one picture.
 While for most model human, ``Truck'' is a kind of ``common sense''.
 
Large pre-trained models have shown their potential in providing such ``common sense'' knowledge to numerous downstream vision and language tasks \cite{kenton2019bert,dosovitskiy2020image}.
The vision-language pre-trained models such as CLIP-like models \cite{zhong2022regionclip,zhou2022coop,zhou2022conditional}  are good candidates to tackle CZSL task, as  CZSL is a multi-modal task involving both texts and images. 
Despite their effectiveness, it is infeasible to fine-tune these large pre-trained model due to their massive scale. 
To utilize these models efficiently, ``prompt tuning'' is proposed to adapt the pre-trained models for the downstream tasks \cite{brown2020language,gao2021making,lester2021power,jiang2020can,shin2020autoprompt}.
Although extensive studies have been conduct in both visual and text prompt tuning, few attempts have been applied to the multi-modal prompt tuning.

In this paper, we proposed a framework named Multi-Modal
Prompt Tuning (MMPT) to explore the potential of applying pre-trained vision-language model to CZSL task. 
As shown in Figure \ref{fig:framework}, we designed a three-branch architecture to align the CLIP-like model CoOp \cite{zhou2022coop} to CZSL task. 
The inputs of the attribute branch and object branch are sentences encoding the attribute or object information (e.g., 'a photo of a  $<$object$>$').
The vision encoder will extract the visual features from the given images. 
Instead of using prompts in single modality,
we proposed shared-prompts across language and vision encoders to bring the gaps between the different modalities.
Also, we add visual patch prompts to further improve the generalization ability for MMPT. 
Unlike other downstream tasks, our experimental results demonstrate OW-CZSL is a tough task that vanilla CLIP-like models can not easily beat current state-of-the-arts.  
And our  MMPT achieves significant  improvement over current SOTA methods, boosting the AUC score from $26.6$ to $29.8$ on the UT-Zappos dataset, and pushing the best AUC score on the MIT-states dataset to $4.1$, which is 150\% better than the current best score $1.6$\cite{mancini2021open}.

The rest of the paper is organized as follows. Section \ref{sec:relatedWork}  reviews related work. Section \ref{sec:model} introduces the MMPT model. Section \ref{sec:exp} presents our experimental results.  Section \ref{sec:conclusion} concludes the paper.




\section{Related Work}
\label{sec:relatedWork}

\subsection{Prompt Learning}
Prompt learning is first introduced in NLP community \cite{gao2021making,lester2021power,jiang2020can,shin2020autoprompt}, which attaches instructions to the input data of the downstream tasks so that the large pre-trained LM can better ``understand'' the downstream tasks such as image-text retrieval, visual question answering, visual grounding. Through carefully designed prompt template, pre-trained LM achieves considerable performance without additional training. Instead of manual prompt engineering, Li and Liang \cite{li-liang-2021-prefix} works on learning continuous prompts. Lester \etal proposes end-to-end soft prompt learning approach \cite{lester2021power} with various design options including prompt length, prompt initialization method, LM adaptation steps, prompt ensembling method is introduced to boost the performance.

\subsection{Vision Language models}
Large-scale visual language models (VLMs) have been developed to utilize rich knowledge among multiple modalities. CLIP \cite{radford2021learning} introduces contrastive learning framework between language and images to do image classification, through pairwise cosine similarity between text prompt ``a photo of a [class]'' and the given image, the model is then pretrained on 400 million curated image-text pairs. ALIGN \cite{jia2021scaling} is similar to CLIP but utilizes a larger noisy image-text dataset for pretraining. Instead of whole image-text matching, RegionCLIP \cite{zhong2022regionclip} proposes to do region-text matching, off-the-shelf region proposal network (RPN) is used to generate image patch, and a CLIP-style prompt template is used to inject text modality. To efficiently tune the pretrained VLMs for downstream tasks, CoOp \cite{zhou2022coop} is proposed to insert a learnable prompt token to the feature embedding space. CoCoOp \cite{zhou2022learning-coop} further adapts the unified prompt towards each input instance to improve the performance on unseen class. ALPRO \cite{li2022align} introduces standalone prompter module to generates soft pseudo labels to identify entities in video, independent transformers are used to encode video and text, instance-level alignment is then learnt using video-text contrastive loss. Furthermore, Li \etal \cite{li2022clip} designed event prompt functions to encode events into natural language sentences.

\subsection{Compositional Zero-shot Learning}
EpiCA \cite{xu-etal-2021-zero} utilizes an episode-based cross-attention network, which contains inductive to learn from labeled pairs and transductive training phase to update the model with unlabeled testing set. Li \etal \cite{li2022siamese} build individual prototypes for attributes and objects, a state transition module is proposed to produce virtual compositions thus improving the model robustness. To emphasize the independence between attributes and objects, Frank \etal \cite{ruis2021independent} proposed ProtoProp, a prototype propagation graph method, the model first generates independent object prototype, a compositional prototypical representation is then learnt for novel attribute-object pairs. Li \etal \cite{li2020symmetry} consider the task in the view of symmetry learning, various losses are computed based on group symmetry properties. CompCos \cite{mancini2021open} estimates the feasibility of each composition, with the assumption that similar states can be assigned to similar objects. CGE \cite{naeem2021learning} uses graph structure to inject the dependency between states, objects and compositions. KG-SP model \cite{karthik2022kg-sp} uses two independent classifiers to predict object category and attributes separately, external knowledge is added to discard unfeasible pairs.

Among the existing CZSL works, most of them focus on closed-world setting, in which the search space for the model output is the number of compositional pairs occurred in the testing set.
This is an unrealistic hypothesis with a relatively small number of composition categories. While CompCos \cite{mancini2021open} and KG-SP \cite{dosovitskiy2020image} consider a more challenging open-world scenario (OW-CZSL), where the actual number of object-attribute pairs in the testing set is unknown and the output space of the model is total number of combinations of all possible objects and attributes ($\text{Number of Obj.} \times \text{Number of Attr.} $). 
In our work, we target on the more challenging scenario OW-CZSL.
\section{Method} 
\label{sec:model}

\subsection{Problem Definition}
The goal of compositional zero-shot learning (CZSL) is to recognize the composition of multiple concepts (objects and attributes) from a given image. 
Specifically, each sample in the training set 
$\mathcal{T}_{train}=\{(x,c)\}_{i=1}^{M}$ contains an image sample $x$ and a composition $c$, where $x \in \mathcal{X}$ is an element in image space  $\mathcal{X}$ and $c=(a,o)$ is a composition label consisted of attribute $a$ and objective $o$.

Given a set of attribute labels $\mathcal{A}$   and a set of object labels $\mathcal{O}$,  the complete composition set $\mathcal{C}$ is constructed as 
$\mathcal{C}= \mathcal{A} \times \mathcal{O}= \{(a,o)|a \in \mathcal{A}, o \in \mathcal{O}\}$.
All the compositions  appearing in training set form the \textit{seen} set $\mathcal{C}^s$, a subset of complete composition set $\mathcal{C}$. 
While the set including compositions never seen in training set is called  \textit{unseen} set $\mathcal{C}^u$. 
The difficulty of CZSL task can be quite different, depending on the output space of the prediction \cite{xian2019zero}.
Given a new test image $x$,  its prediction through a model is denoted as $\hat c$, 
the learning scenarios can be categorized as follows:
\begin{itemize}[noitemsep,topsep=0pt]
    \item supervised learning: $\hat c \in \mathcal{C}^s$,
    \item zero-shot learning: $\hat c \in \mathcal{C}^u$,
    \item generalized zero-shot learning: $\hat c \in \mathcal{C}^s \cup  \mathcal{C}^u $,
    \item open world zero-shot learning: $\hat c \in \mathcal{C}$.
\end{itemize}
In the standard zero-shot learning setting, only the unseen compositions $\mathcal{C}^u$ are predicted during the testing \cite{misra2017red}.
Most recent CZSL work consider the generalized scenario \cite{purushwalkam2019task,li2020symmetry}, 
 where the testing samples can come from either seen set $\mathcal{C}^s$ or unseen set $\mathcal{C}^u$.
 Compared to the standard zero-shot learning, this scenario is more challenging since the model will naturally be biased to the seen compositions.
The most challenging case is the open world zero-shot learning (OW-CZSL) \cite{mancini2021open,karthik2022kg-sp},
where the testing samples can be drawn from the whole composition set $\mathcal{C}$.
In this case, the output space is so large that can hardly be generalized from the small amount of seen compositions ($|\mathcal{C}| \gg  |\mathcal{C}^s|$). 
Our focus is tackling the most challenging OW-CZSL scenario that has no prior assumptions on the unseen set during testing. 

\subsection{Prompt tuning for Compositional Learning}
Pre-trained vision-language models such as CLIP\cite{radford2021learning}, CoOP\cite{zhou2022learning-coop}, and Co-CoOP \cite{zhou2022conditional} have shown promising performance on various downstream tasks.
We are the first to explore the potential of CLIP-like models on the CZSL task. 

Different from  CLIP and CoOP that only contain text prompt, our proposed Multi-Modal Prompt Tuning (MMPT) consists of both text prompts and vision prompts. 
Specifically, MMPT has three branches, one for vision and two for texts (objective branch and attribute branch). 
A plain Vision Transformer (ViT) \cite{dosovitskiy2020image} is adopted in our vision branch  as our backbone, and both attribute and object branch use Language Transformer \cite{vaswani2017attention} as the encoder. 

To enable the multi-modal prompt tuning, we introduce shared  prompts to the input space at front layers of the vision and text encoders.
For the $(i)$-th layer (no matter vision or text branch), the input learnable prompt is denoted as
$t_{i} \in \mathbb{R}^d$, which is a vector with dimension $d$.
The word embeddings $W$ in text encoders (attribute and object branch) are $d_l$-dimentional vectors, and the dimension of image embedding $E$ in vision encoder is denoted as $d_v$. 
To share the mutual prompt $t_i$ to multiple modalities, 
projector function $g_v(t_i):\mathbb{R}^d \rightarrow{\mathbb{R}^{d_v}}$ is used to map shared prompt  $t_i$ to the vision branch.
Similarly, function $g_o(t_i):\mathbb{R}^d \rightarrow{\mathbb{R}^{d_l}}$ and $g_a(t_i):\mathbb{R}^d \rightarrow{\mathbb{R}^{d_l}}$ will project share prompt into learnable parameters in object and attribute branch respectively. 


\begin{table*}[!htp]\centering
\caption{Dataset details.}
\label{tab:datasets}
\begin{tabular}{l|cc|cc|ccc|ccc}\toprule
&  & & \multicolumn{2}{c} {Training} & \multicolumn{3}{c}{Validation} & \multicolumn{3}{c}{Test} \\\cline{4-11}
Dataset & Obj. & Attr.  & Seen pair & Img. &Seen pair &Unseen pair & Img. & Seen pair & Unseen pair & Img. \\
\midrule
UT-Zappos \cite{yu2014fine} & 12 & 16  & 83 & 23k & 15 & 15 & 3k & 18  & 18   & 3k \\
MIT-States \cite{isola2015discovering} & 245 & 115 & 1262 & 30k  & 300 & 300 & 10k & 400 & 400   & 13k \\
\bottomrule
\end{tabular}
\end{table*}

In the input layer of our vision encoder, similar to ViT \cite{dosovitskiy2020image} and VPT \cite{jia2022vpt}, 
 the  $j$-th patch $x^{(j)}$ of an image $x$ is embedded into a $d_v$-dimensional vector.
 Additionally, we add visual prompt \cite{bahng2022exploring},  which is a learnable vector parameterized by $\phi$.
 Then the initial embeddings in the vision branch can be formulated as:
 \begin{equation}
        E_0=\{e_0^j\}_{j=1}^K =\{Embedding(x^{(j)}+{\phi})\}_{j=1}^K,
        \label{equ:visial_promt}
 \end{equation}
 where  $K$ is the number of patches in the image. 
 Different from ViT\cite{dosovitskiy2020image} and VPT \cite{jia2022vpt}, we inject learnable shared prompt $g_v(t_i)$ to the input space of the $i$-th layer, together with learned embedding $E_{i-1} =\{e_{i-1}^j \in \mathbb{R}^{d_v}\}_{j=1}^K$  from preceding layer and the hidden feature $z_{i-1}$ that encodes the learnable attribute token and object token. 
The above-mentioned forward process is formalized as follows: 
 \begin{equation}
     [E_i,z_i]=L_i^{(v)}([g_v(t_i),E_{i-1},z_{i-1}]) \:, 1 \leqslant i \leqslant h_s, 
     \label{equ:vision_front_forward}
 \end{equation}
 where $[\cdot,\cdot]$ indicates the concatenation operation on vectors, $L_i^{(v)}$ denotes the $i$-th layer of vision Transformer.
Note that, only the first $h_s$ layers have 
the layer-specific shared prompts, the subsequent layers starting from $h_s+1$ will just process the prompts from the previous layers:
  \begin{equation}
[G^{(v)}_j,E_i,z_j]=L_j^{(v)}([G^{(v)}_{j-1},E_{j-1},z_{j-1}]) \:, h_s < j \leqslant h_v,
\label{equ:vision_tail_forward}
 \end{equation}
where $h_v$ is the number of total layers in the vision Transformer. 
As $j$ starts from $h_s+1$, the initial value of $G^{(v)}_{h_s}$ is equal to $g_v(t_{h_s})$.
As shown in Section \ref{sec:hyper_parameter}, it is unnecessary to insert layer-specific shared prompts to all the Transformer blocks. 


We also project the shared prompt $t_i$ to language Transformers that processing through mapping function $g_a(t_i)$ and $g_o(t_i)$. 
The initial layer of language Transformer in attribute branch will generate word embeddings $W_0^{(a)} \in \mathbb{R}^{d_l}$ from text. 
At each following layer $i$,
the inputs include 
the layer-specific prompt $g_a(t_{i})$, the fixed embeddings $W_{i-1}$ and the attribute token $y_{i-1}^{(a)}$  from previous layer: 
   \begin{equation}
       [W_i^{(a)},y_i^{(a)}]=L_i^{(a)}([g_a(t_i),W_{i-1}^{(a)},y_{i-1}^{(a)}]) \:, 1 \leqslant i \leqslant h_s.
       \label{equ:att_front_forward}
   \end{equation}
Similar to vision branch forward process shown in Equation (\ref{equ:vision_front_forward}) and (\ref{equ:vision_tail_forward}),
language Transformer layers after the $h_s$-th layer will reuse the prompts learned from the previous layer as their input prompts:
   \begin{equation}
[G^{(a)}_{j},W_i^{(a)},y_i^{(a)}]=L_i^{(a)}([G^{(a)}_{j-1},W_{j-1}^{(a)},y_{j-1}^{(a)}]) \:, h_s < j \leqslant h_a,
 \label{equ:att_tail_forward}
\end{equation}
where $h_a$ is the number of layers in Transformer of attribute branch, and $G_{h_s}^{(a)}=g_o(t_{h_s})$ is the prompts directly forwarded across the tail layers.
The learning process of object branch is the same as that of the attribute branch:
 \begin{align}
   [W_i^{(o)},y_i^{(o)}]&=L_i^{(o)}([g_o(t_i),W_{i-1}^{(o)},y_{i-1}^{(o)}]) \:, 1 \leqslant i \leqslant h_s \\
   [G^{(o)}_{j},W_i^{(o)},y_i^{(o)}]&=L_i^{(o)}([G^{(o)}_{j-1},W_{j-1}^{(o)},y_{j-1}^{(o)}]) \:, h_s < j \leqslant h_o,
\label{equ:forward_obj}
  \end{align} 
  where $h_o$ is the number of layers in Transformer of object branch.
The final representations are taken from the last layers to compute the the prediction probabilities of the attribute and object. 
Taking the prediction of attribute for example, given a specific attribute $a^*$, its probability over image $x$ is computed as follows: 
\begin{equation}
\rho_a(a=a^*,z=z_{h_v})=\frac{\exp(\cos(\omega(y_{h_a}^{(a^*)}),\nu(z_{h_v}))/\tau)}{\sum\limits_{a \in \mathcal{A}}\exp(\cos(\omega(y_{h_a}^{(a)}),\nu(z_{h_v}))/\tau)},
\end{equation}
where $cos(\cdot)$ denotes the cosine similarity, $\omega(\cdot)$ and $\nu(\cdot)$ are projecting functions, $z_{h_v}$ is the output vector from  the $h_v$-th layer of the vision Transformer,  $y_{h_a}^{(a)}$ is the representation of attribute token from the final layer of attribute branch, and $\tau$ is a fixed temperate parameter. 
The score of object prediction can be computed in a similar way as follows:
\begin{equation}
\rho_o(o=o^*,z=z_{h_v})=\frac{\exp(\cos(\omega(y_{h_o}^{(o^*)}),\nu(z_{h_v}))/\tau)}{\sum\limits_{o \in \mathcal{O}}\exp(\cos(\omega(y_{h_o}^{(o)}),\nu(z_{h_v}))/\tau)}.
\end{equation}

Previous work  have shown that predicting attribute and object independently can best utilize the training data \cite{misra2017red,karthik2022kg-sp}. 
In our proposed MMPT, the attribute branch and object branch can be viewed as two separate classifiers, which are combined by the following loss function:
\begin{equation}
\begin{aligned}
\mathcal{L}&=\mathcal{L}_o+\mathcal{L}_a \\
&=-\frac{1}{|\mathcal{T}|} \sum \limits_{\{x,(a,o)\} \in \mathcal{T}}\log\rho_o(o,z)+ \log\rho_a(a,z).
\end{aligned}
\label{equ:loss}
\end{equation}
As shown in Equation (\ref{equ:loss}), during training, both the cross-entropy loss for attribute and object branch are minimized.
In the inference process, we will consider all possible compositions,
and the pair of attribute and object with highest score is taken as the prediction:
\begin{equation}
\hat c = \argmax \limits_{(a,o) \in C} \rho_o(o,z) \rho_a(a,z),
\end{equation}
where $z$ is the vision features extracted from the given image $x$ through the vision encoder.
As our setting is OW-CZSL, the searching space of outputs is the full composition set $C$, where most the compositions are never seen in the training data. 
\section{Experiments}
\label{sec:exp}

 \begin{table*}[!htp]\centering
\caption{Open World CZSL results on UT-Zappos and MIT-states. The  best seen (S) and unseen accuracy (U), best harmonic mean (HM), and area under the curve (AUC) are reported on the test sets of the two datasets.}
\label{tab:main}
\begin{tabular}{l|rrrr|rrrrr}\toprule
\
&\multicolumn{4}{c}{UT-Zappos Test} &\multicolumn{4}{c}{MIT-states test} \\\cmidrule{2-9}
&S &U &HM &AUC &S &U &HM &AUC \\\midrule
LE+ \cite{misra2017red} &60.4 &36.5 &30.5 &16.3 &14.2 &2.5 &2.7 &0.3 \\
VisProd \cite{misra2017red} &54.6 &42.8 &36.9 &19.7 &20.9 &5.8 &5.6 &0.7 \\
SymNet \cite{li2020symmetry} &53.3 &44.6 &34.5 &18.5 &21.4 &7 &5.8 &0.8 \\
AoP \cite{nagarajan2018attributes} &50.9 &34.2 &29.4 &13.7 &16.6 &5.7 &4.7 &0.7 \\
TMN \cite{purushwalkam2019task} &55.9 &18.1 &21.7 &8.4 &12.6 &0.9 &1.2 &0.1 \\
CompCos \cite{mancini2021open} &59.3 &46.8 &36.9 &21.3 &25.4 &10 &8.9 &1.6 \\
KG-SP \cite{karthik2022kg-sp} &61.8 &52.1 &42.3 &26.5 &28.4 &7.5 &7.4 &1.3 \\ \hline
CoOp \cite{zhou2022coop} &{{54.7}}	&{{45.7}}	& {35.9}	& {20.0} & {27.9}  & {16.9}  & {13.6}  & {3.3 }\\

MMPT(Ours) &\textbf{63.3} &\textbf{56.0} &\textbf{45.1} &\textbf{29.8} & \textbf{32.6} &\textbf{17.7} &\textbf{15.0} &\textbf{4.1} \\

\bottomrule
\end{tabular}
\end{table*}

\subsection{Datasets}
\label{sec:dataset}
We evaluated our proposed MMPT and baselines with two standard 
datasets for Compositional Zero-Shot Learning: UT-Zappos \cite{yu2014fine}  and MIT-States \cite{isola2015discovering}.
UT-Zappos contains around 33k images of shoes, labelled by 12 types (objects) and 16 materials (attributes).
MIT-States contains 53k images belonging to 245 object categories and 115 attributes.
Although the numbers of images in the two datasets are close, 
the output space of MIT-States (28,175 possible compositions) is much larger than that of  UT-Zappos (192 compositions).
Therefore, the AUC and HM scores on MIT-States are much lower than the scores on UT-Zappos, as most of the compositions in MIT-States are unseen to the models. 

We follow the standard splits of dataset \cite{karthik2022kg-sp,purushwalkam2019task}.
The detailed information can be found in Table \ref{tab:datasets}. 
Note that, the output space of OW-CZSL setting is far more larger than that of the generalized zero-shot learning, which is the evaluation setting adopted by most previous CZSL models \cite{li2020symmetry,li2022siamese,xu-etal-2021-zero}.
Taking MIT-States dataset for example, 
the output searching space of traditional CZSL setting will contain 1,262 seen compositions and 400 unseen compositions. 
While in the OW-CZSL setting, all 28,175 possible compositions are considered in the output space, where the most of compositions (26,114 out of 28,175) are not included in any splits of the datasets.

\subsection{Metrics and Baselines}
Following the setting of previous work \cite{li2022siamese,karthik2022kg-sp,purushwalkam2019task}, 
the following metrics are used in our evaluations: 
1) Seen (S) is the metric that measures the accuracy on the compositions that are already seen in the training set. This metric can reflect the performance of model in supervised learning. 
2) Unseen (U) is the prediction accuracy tested only on unseen compositions, which is a primary indicator of the model's generalization ability.  
3) Harmonic Mean (HM) is an average score that balances the seen and unseen accuracy.
4) Area Under the Curve (AUC) is the overall measurement for the generalized CZSL methods, which reflects the trade-off between seen and unseen set at various calibration biases. 

We compare our model with the existing methods on compositional zero-shot learning task. The previous CZSL models LE+ \cite{misra2017red}, VisProd\cite{misra2017red}, SymNet \cite{li2020symmetry}, AoP \cite{nagarajan2018attributes}, TMN \cite{purushwalkam2019task} are originally designed for closed-world setting, while CompCos \cite{mancini2021open}, KG-SP \cite{karthik2022kg-sp} and our proposed MMPT are in open-world setting. 
We train these models with their default settings, and report their results on the open world CZSL setting. 
More details can be found in the Appendix.


\begin{figure} [htbp]
  \raggedleft
   \includegraphics[width=\columnwidth]{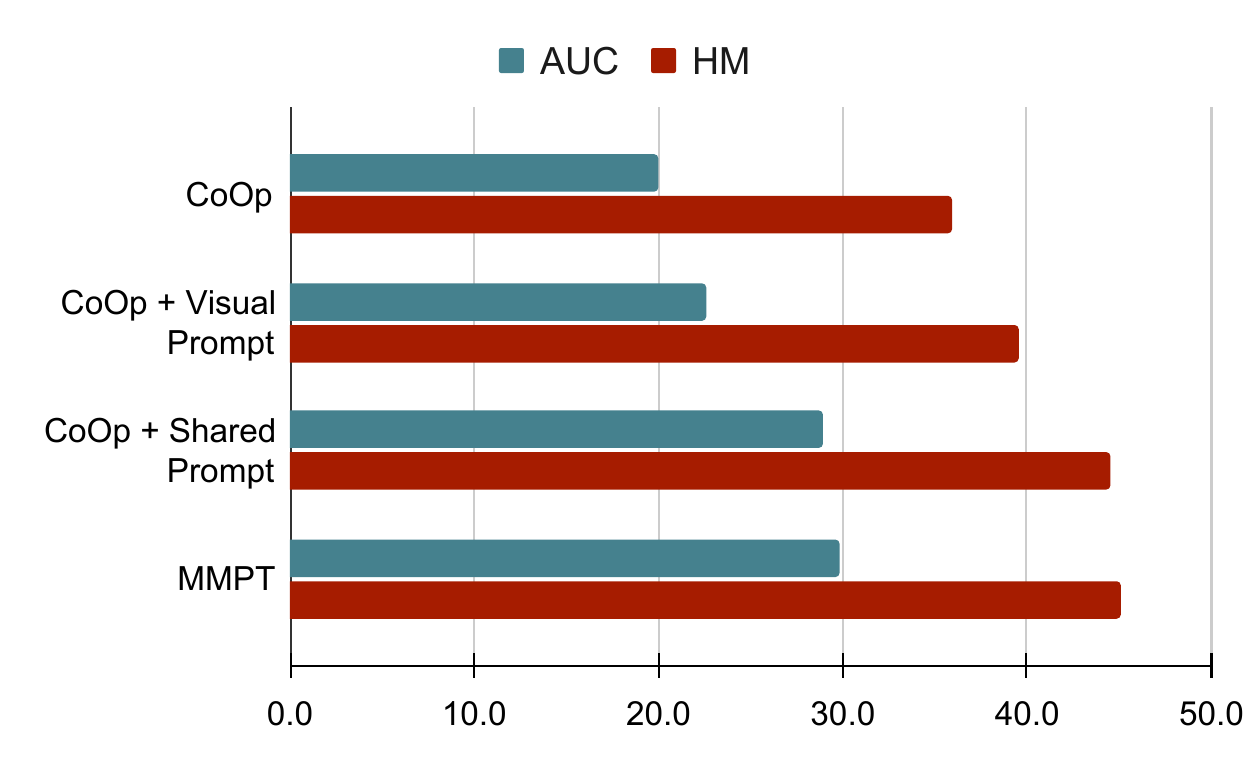}
       \caption{Comparison of AUC and HM scores on UT-Zappos for  CoOp,  CoOp + Visual Prompt, CoOp + Shared Prompt and our proposed MMPT. 
       }
       \label{fig:ablation}
\end{figure}

\begin{table}[!htp]\centering
\caption{Comparison of seen and unseen accuracy scores on UT-Zappos for CoOp,  CoOp + Visual Prompt, CoOp + Shared Prompt and our proposed MMPT. }\label{tab:ablation}
\begin{tabular}{l|rr|rrr}\toprule
Model &S &$\Delta S$ &U &$\Delta U$ \\\midrule
CoOp &54.7 &0.0 &45.7 &0.0 \\
CoOp + Visual Prompt &58.0 &3.3 &47.3 &1.6 \\
CoOp + Shared Prompt &\textbf{63.8} &9.1 &{54.5} &8.8 \\
MMPT &63.3 &8.6 &\textbf{56.0} &10.3 \\
\bottomrule
\end{tabular}
\end{table}

\subsection{Quantitative Results}

A good CZSL model should be ``smart'' enough to correctly connect the attribute and object  when there are sufficient amount of seen compositions.
Also, the model needs to be ``knowledgeable'' to foresee the unseen compositions, when the majority of output space is unknown. 
The results of open world CZSL are reported in Table \ref{tab:main}.

The first seven rows are results of state-of-the-arts in CZSL.
Besides, we also add results of applying CoOp \cite{zhou2022learning-coop} on CZSL task.
On both MIT-States dataset and UT-Zappos dataset, our MMPT achieves significant performance improvements over these baselines in terms of all metrics, which indicates MMPT is a  ``smart'' and ``knowledgeable'' CZSL model.
Specifically, MMPT boosts the AUC score from $26.5$ (obtained by KS-SP \cite{karthik2022kg-sp}) to $29.8$ on UT-Zappos dataset, 
and on MIT-states dataset MMPT achieves an amazing AUC score up to $4.1$, which is 150\% better than the current state-of-the-art CompCos \cite{mancini2021open}.

To our surprise, the best runner up on UT-Zappos dataset is KG-SP\cite{karthik2022kg-sp}, a relatively light-weight model, instead of the recent ``rock star '' CoOp \cite{zhou2022learning-coop}, which has shown great performance in various downstream tasks. 
This phenomenon confirms that open world CZSL is a challenging task that can not be easily brute forced by large pre-trained vision-language models. 
On the other hand,  the pre-trained models have their own strengths in the scenarios where most the compositions are unseen, since they inherently contain external knowledge. 
On MIT-States, CoOp \cite{zhou2022learning-coop} baseline boosts the AUC score to $3.3$, while the highest score of previous CZSL methods is only $1.6$ (obtained by CompCos \cite{mancini2021open}). 
Previous work also notice the importance of being ``knowledgeable''. 
CompCos \cite{mancini2021open} learns cosine similarities between visual and compositional embeddings, while KG-SP \cite{karthik2022kg-sp} brings external knowledge to CZSL model through ConceptNet \cite{speer2017conceptnet}. 
CLIP-like models such as CoOp \cite{zhou2022learning-coop} are pre-trained on large scale data by learning the similarities between images and texts, which can be viewed as the combined effects of KG-SP \cite{karthik2022kg-sp} and CompCos \cite{mancini2021open} with broader vision. 
Standing on the shoulder of CoOp \cite{zhou2022learning-coop}, our proposed MMPT is unsurprisingly more ``knowledgeable'' than the models specially designed for CZSL.

\begin{figure*} [t]
  \begin{subfigure}{.49\textwidth}
   \centering
\includegraphics[height=4.1cm]{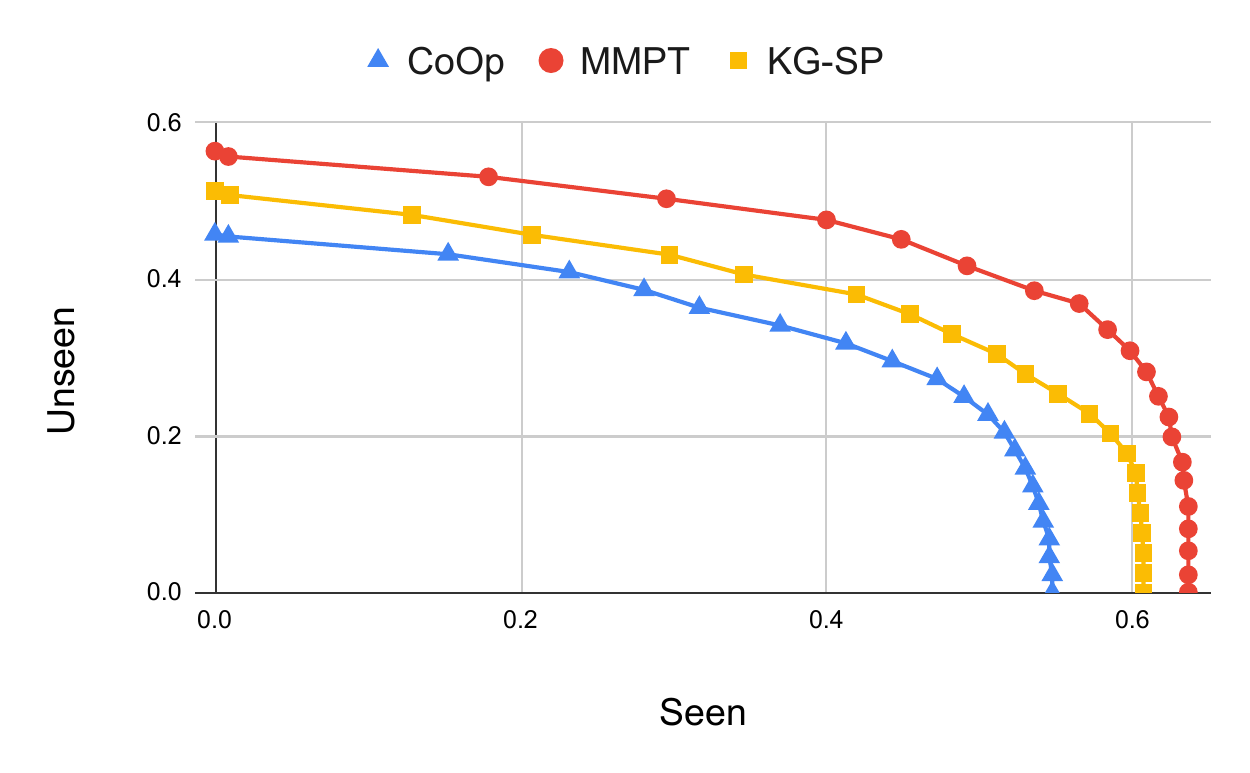}
       \caption{UT-Zappos}
       \label{fig:bias-UT}
       \end{subfigure}
         \begin{subfigure}{.49\textwidth}
  \centering
\includegraphics[height=4.1cm]{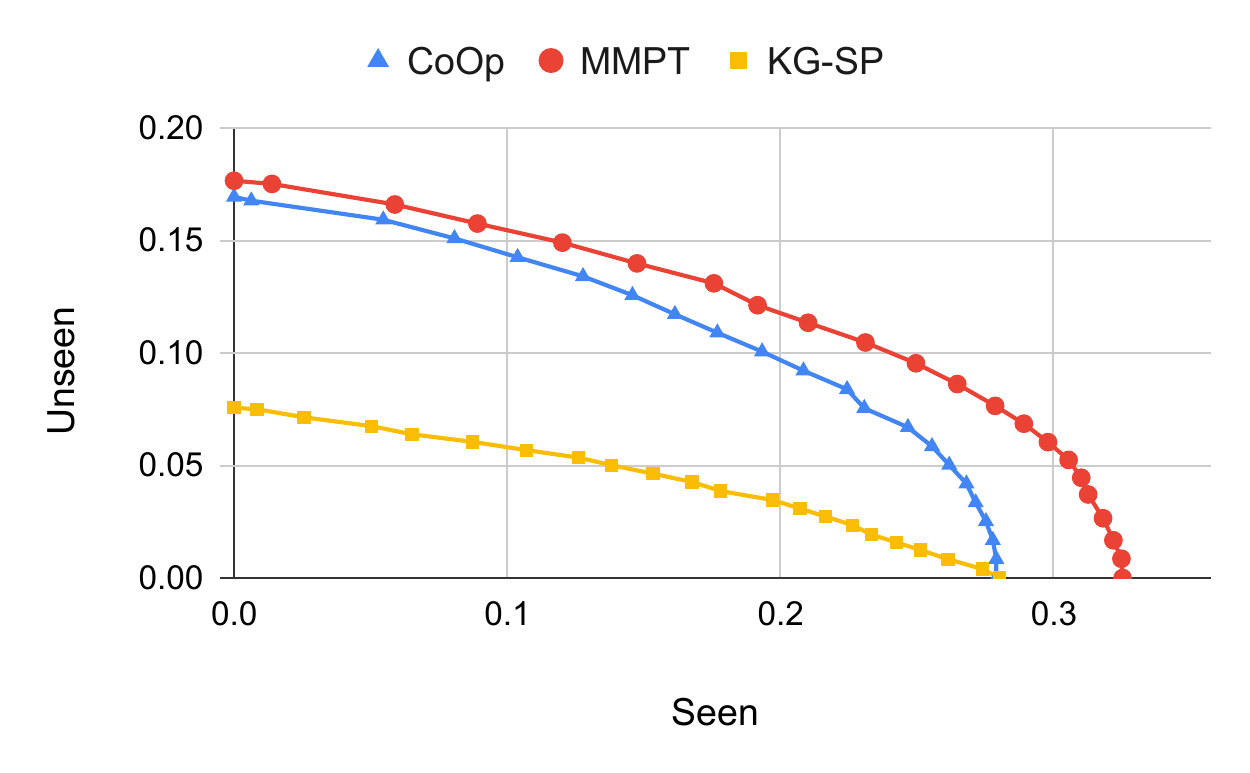}
       \caption{MIT-States}
       \label{fig:bias-MIT}
       \end{subfigure}
\caption{Unseen-seen accuracy on UT-Zappos dataset and MIT-States dataset under various calibration biases.
}
\label{fig:bias}
\end{figure*}
\subsection{Trade-off between seen and unseen accuracy}

Figure \ref{fig:bias} plots the unseen-seen accuracy curves on the UT-Zappos and MIT-States dataset for CoOp baseline \cite{zhou2022coop}, the runner-up method KG-SP \cite{karthik2022kg-sp}, and our our proposed MMPT. For all the models,  the seen accuracy decreases while the unseen accuracy increases, as the calibration value increases. 
It is known as an important trade-off when models are robust for interventions. 

On UT-Zappos dataset, the unseen-seen curve of KG-SP stands in between of CoOp\cite{zhou2022coop} and MMPT, which is consistent with the what we obtained from Table \ref{tab:main}. 
On  MIT-States dataset, the patterns shown in Figure \ref{fig:bias-MIT} confirm that CoOp\cite{zhou2022coop} and MMPT are more ``knowledgeable'' than KG-SP \cite{karthik2022kg-sp} that their values of unseen accuracy are much larger than that of KG-SP \cite{karthik2022kg-sp} under same level of seen accuracy.

\subsection{Ablation Study}
\label{sec:Ablation}
In this section, we reveal secrets why MMPT is ``smarter'' than CLIP-like baseline  CoOp \cite{zhou2022learning-coop}. 
Compared to CoOp \cite{zhou2022learning-coop}, MMPT contains two novel components:
the visual prompts described in Equation (\ref{equ:visial_promt}), 
and the shared multimodal prompt shown from Equation (\ref{equ:vision_front_forward}) to (\ref{equ:att_tail_forward}).
As can be seen from Figure \ref{fig:ablation},
both Visual Prompt and Shared Prompt help to improve the performance of vanilla CoOp \cite{zhou2022learning-coop}. 
Specifically, visual prompt alone can boost the original AUC from $20.0$ to $22.6$, while multi-modal shared prompt achieves a more remarkable improvement, increasing the AUC score to $28.9$.
With the combined usage of visual prompt and multi-modal shared prompt, it is not surprised that our MMPT achieve the best performance. 
Referring to the margin scores of seen and unseen accuracy in Table \ref{tab:ablation},
MMPT has a slightly smaller seen score along with a much better unseen accuracy, compared to baseline ``CoOp + Shared Prompt''.
This phenomenon indicates the combination of both visual prompt and shared prompt is beneficial, which helps to prevent the overfitting (decreasing the seen accuracy) without hurting the generalization ability (increasing the unseen accuracy).

\subsection{Hyper-parameter Analysis}
\label{sec:hyper_parameter}
There are two important hyper-parameters that may impact the performance of MMPT,
prompt length $d$ for shared prompt and the depth of layers with layer-specific prompts denoted by $h_s$ in Equation (\ref{equ:vision_front_forward}) to (\ref{equ:att_tail_forward}). 
Due to page limitation, we only present the results of the UT-Zappos, the results of MIT-states can be found in Appdenix.

\begin{figure} [htbp]
  \raggedleft
   \includegraphics[width=\columnwidth]{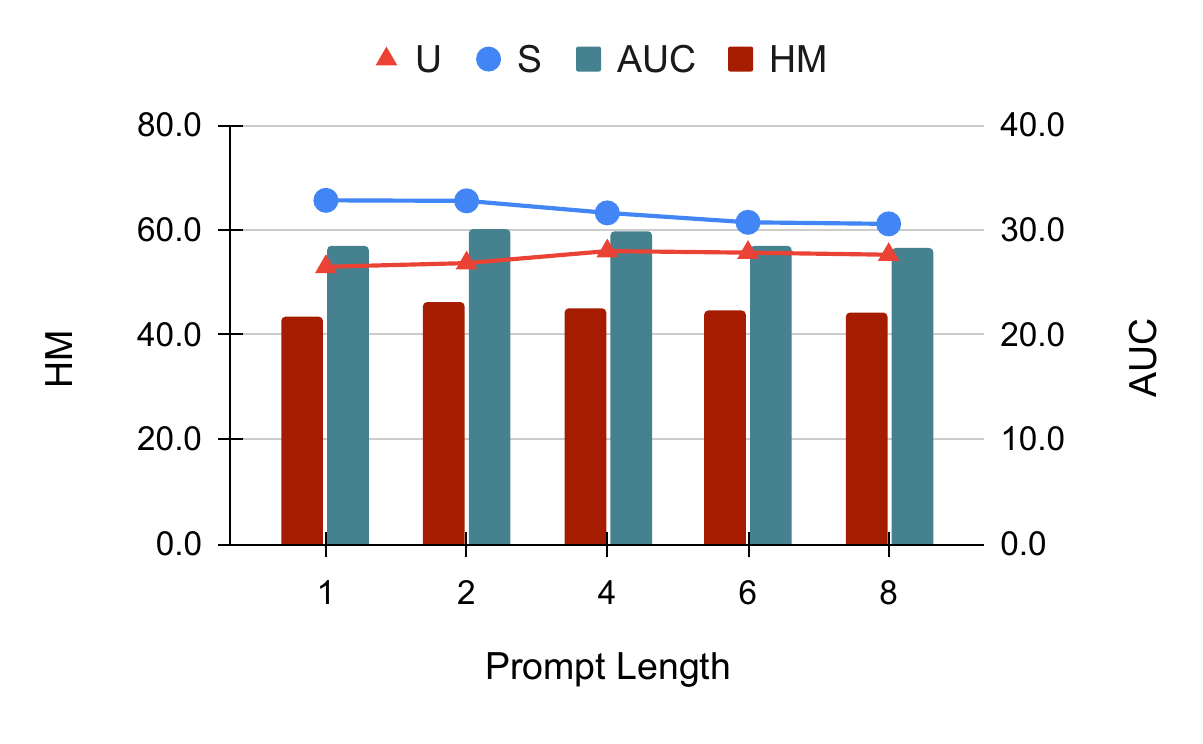}
       \caption{Effect of prompt length on UT-Zappos. 
       }
       \label{fig:prompt_length}
\end{figure}
\begin{figure} [htbp]
  \raggedleft
   \includegraphics[width=\columnwidth]{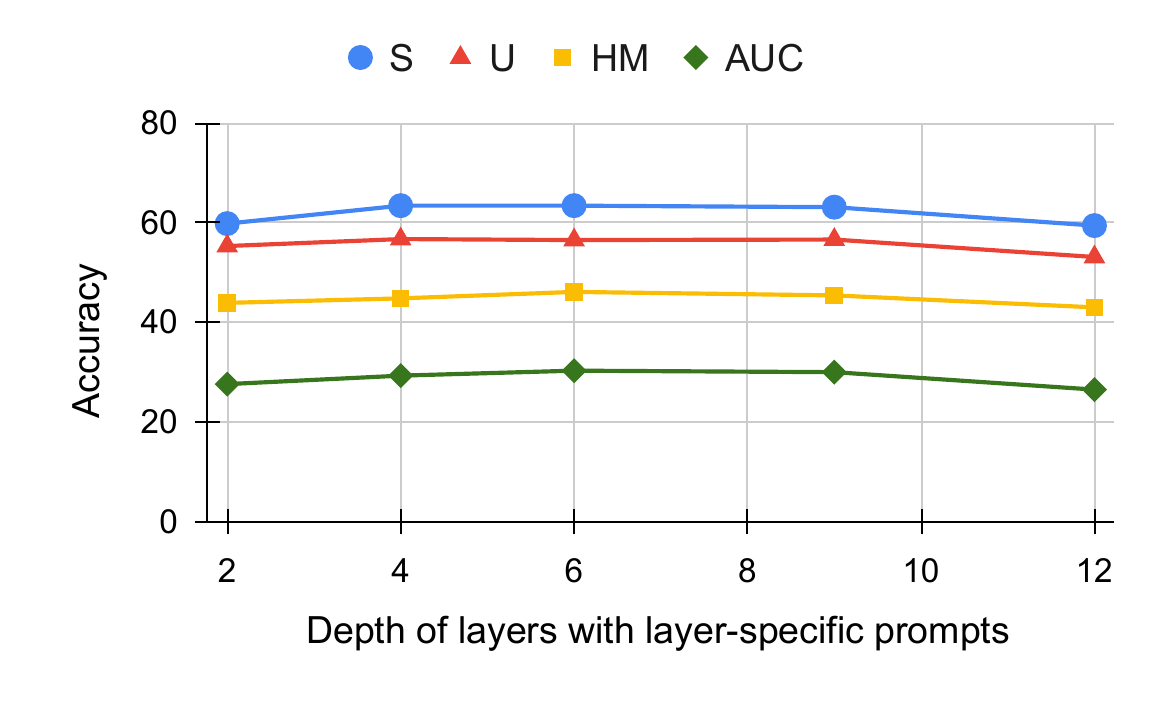}
       \caption{Effect of  the depth of layers with the independently learnable layer-specific prompt on UT-zappos. 
       }
       \label{fig:prompt_depth}
\end{figure}
\noindent\textbf{Shared prompt length}. 
Figure \ref{fig:prompt_length} shows the effect of shared prompt length $d$.  
With the growth of prompt length, the unseen scores increase and the seen accuracy decreases, while the overall performance metrics AUC and HM remain almost unchanged after prompt length $6$, which is set as our hyper-parameter $d$ for UT-Zappos dataset. 
 \\ \hspace*{\fill} \\
\noindent\textbf{Depth of layers with layer-specific prompts}.
Figure \ref{fig:prompt_depth} illustrates the effect of 
the depth $h_s$ which denotes the number of layers with independently learnable layer-specific prompts.
Overall, this hyper-parameter is less sensitive than the above-mentioned prompt length $d$. 
As depth $h_s$ increases, performance first improves followed by slight drops, and the peak values appear in the range from $6$ to $9$. 
This is consistent with the observations in previous work that the front layers of deep neural networks will learn more independent features compared to the tail layers.

\section{Conclusion}
\label{sec:conclusion}
In this paper, we proposed a framework named Multi-Modal
Prompt Tuning (MMPT) to tackle the challenging Open World Compositional Zero-Shot Learning (OW-CZSL)  task.
Different from other downstream vision and language tasks, OW-CZSL is a very challenging task that the vanilla vision-language model CoOp fails to beat current SOTA methods designed for CZSL.  
While extensive experimental results demonstrate that our proposed MMPT is a ``smart'' and ``knowledgeable'' CZSL learner, boosting the state-of-the-art performance by large margins on both UT-Zappos dataset and MIT-States dataset.
Built on the basis of vision-language model CoOp, our MMPT is naturally more ``knowledgeable'' than other CZSL models, and our ablation studies reveal the secret of our ``smartness'' is the combination of using visual prompts and multi-modal shared prompts across different branches.

\clearpage


\bibliographystyle{ieee_fullname}
\bibliography{egbib}

\clearpage
\section{Appendix}
\subsection{Implementation details}

The original CLIP-like model CoOp \cite{zhou2022coop} can not be applied to the CZSL task directly. 
In our implementation, we constructed two text prompt learners for object class names and attribute names, respectively. The image is fed into the visual encoder to extract the visual features, and text prompts are fed into the text encoder to generate language embeddings for object class and attribute class. The final predictions are obtained by selecting the classes with the highest cosine similarity between image feature and corresponding language feature. 

For MMPT, given an input image with the size ${224 \times 224 }$, we randomly select one location within the image and add a visual patch prompt $\phi$ of size ${16 \times 16}$ to the original image, which is randomly initialized using standard normal distribution. The values for the shared prompt $t_i$ of each layer $i$ are randomly initialized separately. The dimension of encoded visual feature $d_v$ is 768 and dimension $d_l$ of text feature in attribute branch and object branch is 512. The number of transformer layers in vision branch $h_s$, attribute branch $h_a$ and object branch $h_o$ are all 12. The length $d$ of multi-modal shared prompt is set to be $6$, 
and the number of layers with layer-specific shared prompt $h_s$ is set to be 9.
The searching process of above mentioned hyper-parameters is analyzed in Section \ref{sec:hyper_parameter}.

All the models are trained on a single NVIDIA Tesla V100 GPU. We use Adam optimizer for training, the learning rate is   $ 5 \times 10^{-5} $, batch size is set as 48.

\subsection{Ablation Study on MIT-States}
\begin{figure} [htbp]
  \raggedleft
   \includegraphics[width=\columnwidth]{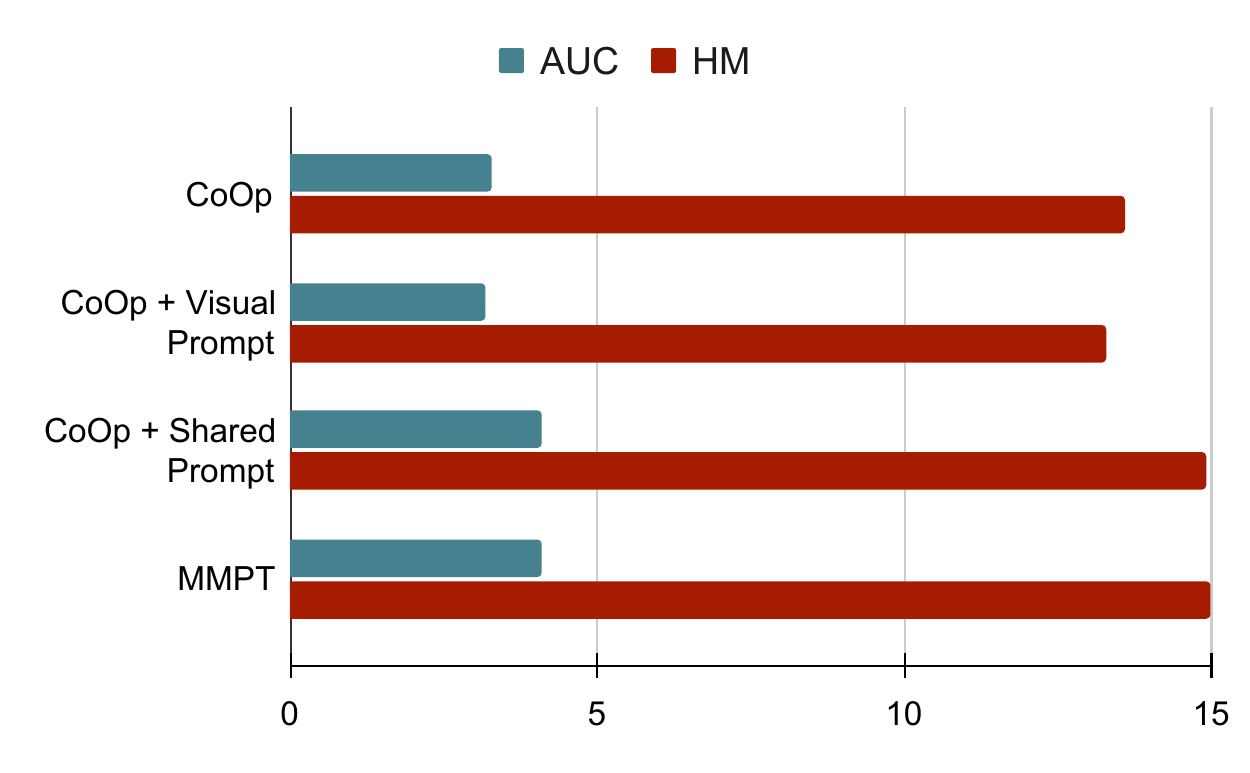}
       \caption{Comparison of AUC and HM scores on MIT-States for  CoOp,  CoOp + Visual Prompt, CoOp + Shared Prompt and our proposed MMPT. 
       }
       \label{fig:ablation-mit}
\end{figure}

\begin{table}[!htp]\centering
\caption{Comparison of seen and unseen accuracy scores on MIT-States for CoOp,  CoOp + Visual Prompt, CoOp + Shared Prompt and our proposed MMPT. }\label{tab:ablation-mit}
\begin{tabular}{l|rr|rrr}\toprule
Model &S &$\Delta S$ &U &$\Delta U$ \\\midrule
CoOp & 27.9	&0.0	&16.9	&0.0 \\
CoOp + Visual Prompt &27.9	&0.0	&16.1	&-0.8 \\
CoOp + Shared Prompt &35.3	&7.4	&16.4	&-0.5 \\
MMPT &32.6	&4.7	&17.7	&0.8 \\
\bottomrule
\end{tabular}
\end{table}

Figure \ref{fig:ablation-mit} and Table \ref{tab:ablation-mit} show the performance comparisons on MIT-States dataset for CoOp\cite{zhou2022coop}, CoOp + Visual Prompt, CoOp + Shared Prompt, and our MMPT. 
The observations from MIT-States dataset shown in Figure \ref{fig:ablation-mit} and Table \ref{tab:ablation-mit} are generally consistent with the analysis in Section \ref{sec:Ablation}.
Compared to vanilla CoOp\cite{zhou2022coop}, most of the performance gain of MMPT is obtained  by our shared prompt scheme and visual prompt can help to prevent the overfitting. 

\subsection{Hyper-parameter Analysis on MIT-States}

\begin{figure} [htbp]
  \raggedleft
   \includegraphics[width=\columnwidth]{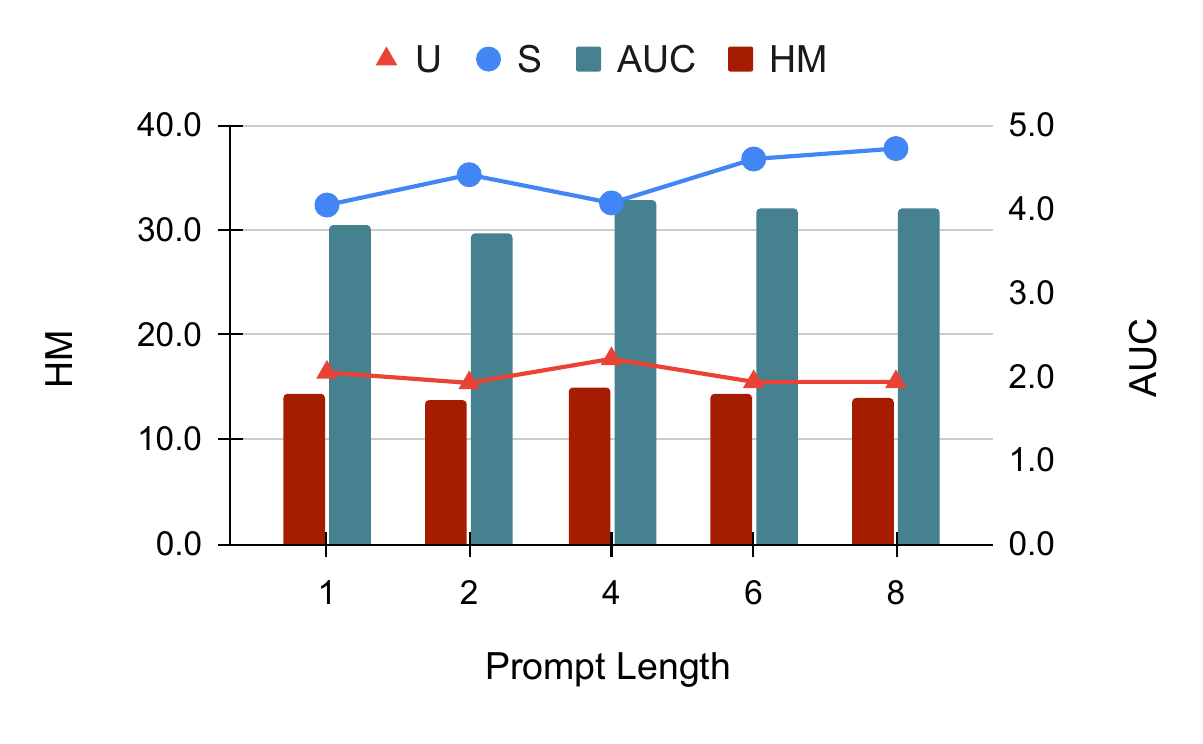}
       \caption{Effect of prompt length on MIT-States dataset. 
       }
       \label{fig:prompt_length_mit}
\end{figure}
\begin{figure} [htbp]
  \raggedleft
   \includegraphics[width=\columnwidth]{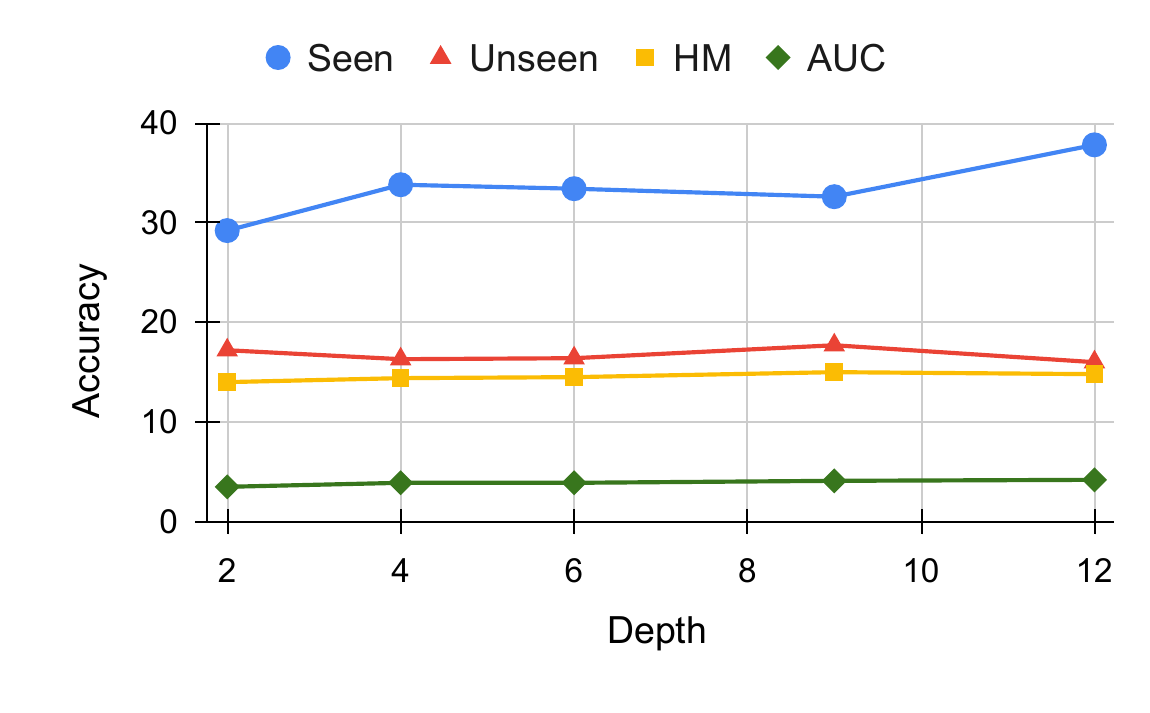}
       \caption{ Effect of the depth of layers with the independently
learnable layer-specific prompt on MIT-States dataset. 
       }
       \label{fig:prompt_depth_mit}
\end{figure}

Figure \ref{fig:prompt_length_mit} shows the effect of shared prompt length $d$ on MIT-States dataset.  
Similar to UT-Zappos dataset,  the overall performance metrics AUC and HM on MIT-States dataset also remain almost unchanged after prompt length $4$, although there are more fluctuations on seen and unseen accuracy on MIT-States dataset.

Figure \ref{fig:prompt_depth_mit} plots the effect of  the depth $h_s$ on MIT-States dataset.
The overall metric AUC and HM generally stay stable, regardless the change of prompt length $d$. 
Similar to the case on UT-Zappos dataset, we also set $d$ to be $9$ on MIT-States dataset.
Compared to Figure \ref{fig:prompt_depth} (depth effect for UT-Zappos dataset), it is more obvious on MIT-States dataset that there is no need to learn independent prompts for all layers.
We can observe that under the same AUC and HM values, learning independent prompts for all layers ($d=12$) will cause a drop in unseen accuracy. 

\end{document}